\documentclass[10pt, a4paper]{article}

\usepackage[final]{lrec2026} 
\usepackage{graphicx}
\usepackage{comment}
\usepackage{multirow}
\usepackage{xcolor}  
\usepackage{pifont}  

\usepackage{listings}
\usepackage{multicol}
\usepackage{tabularx,ragged2e}
\usepackage{booktabs}
\usepackage{float}
\usepackage{comment}

\title{Rewrite the News: Tracing Editorial Reuse Across News Agencies}

\name{
Soveatin Kuntur\textsuperscript{1*},
Nina Smirnova\textsuperscript{2*},
Anna Wroblewska\textsuperscript{1},\\
{\large\textbf{Philipp Mayr\textsuperscript{2},
Sebastijan Razboršek Maček\textsuperscript{3}}}
}

\address{
\textsuperscript{1}Warsaw University of Technology, Poland\\
\textsuperscript{2}GESIS – Leibniz Institute for the Social Sciences, Germany\\
\textsuperscript{3}Slovenian Press Agency, Slovenia\\
\{soveatin.kuntur.dokt, anna.wroblewska1\}@pw.edu.pl,\\
\{nina.smirnova, philipp.mayr\}@gesis.org, sm@sta.si\\
\textsuperscript{*}Equal contribution
}

\abstract{
This paper investigates sentence-level text reuse in multilingual journalism, analyzing where reused content occurs within articles. We present a weakly supervised method for detecting sentence-level cross-lingual reuse \emph{without} requiring full translations, designed to support automated pre-selection to reduce information overload for journalists \cite{holyst2024}. The study compares English-language articles from the Slovenian Press Agency (STA) with reports from 15 foreign agencies (FA) in seven languages, using publication timestamps to retain the earliest likely foreign source for each reused sentence. We analyze 1{,}037 STA and 237{,}551 FA articles from two time windows (October 7–November 2, 2023; February 1–28, 2025) and identify 1{,}087 aligned sentence pairs after filtering to the earliest sources. Reuse occurs in 52\% of STA articles and 1.6\% of FA articles and is predominantly non-literal, involving paraphrase and compositional reuse from multiple sources. Reused content tends to appear in the middle and end of English articles, while leads are more often original, indicating that simple lexical matching overlooks substantial editorial reuse. Compared with prior work focused on monolingual overlap, we (i) detect reuse across languages without requiring full translation, (ii) use publication timing to identify likely sources, and (iii) analyze where reused material is situated within articles. Dataset and code: \url{https://github.com/kunturs/lrec2026-rewrite-news}.
\\ \newline \Keywords{Text reuse detection, multilingual NLP, cross-lingual journalism, computational social science} }

\begin{document}

\maketitleabstract

\section{Introduction}

Text reuse has long been a topic of interest in Natural Language Processing (NLP), from early analyses of classical texts such as the Synoptic Gospels \cite{lee-2007-computational} to applications like plagiarism detection \citep{clough-etal-2002-building} and healthcare communication \citep{dhondt-etal-2016-low}. However, despite its historical significance, broader attention to the phenomenon has declined in recent years. 

Nonetheless, text reuse remains a central practice in multilingual journalism. National news agencies often adapt international reports, paraphrasing, summarizing, or merging multiple foreign sources to produce domestic news articles: factual pieces published on the agency’s newswire and used by other media outlets. Understanding how these articles are constructed sheds light on the invisible processes of news dissemination, editorial decision-making, and cross-lingual information flow, which can help in preparing automated methods for information extraction and the pre‑selection of fragments relevant at the national level.

Analyzing text reuse in this setting uncovers the mechanisms of cross-lingual news dissemination and highlights the need for computational methods capable of tracing multilingual content transformation, essential for transparency and efficiency in global journalism.
This paper investigates how journalistic content is reused and merged across languages within real-world news agency workflows. Specifically, we address the following research questions. For English‑language news articles published by the Slovenian Press Agency (STA)\footnote{\url{https://english.sta.si/}}, we examine: \textbf{(1)} How is content merged from multiple sources? \textbf{(2)} Is text reuse more common in lead paragraphs, the middle of the article or at the end? \textbf{(3)} Which article parts are being reused more frequently?

To answer these questions, we propose a weakly supervised framework for identifying and classifying sentence-level reuse in multilingual newswire corpora. Using data from 16 news agencies in seven languages (Italian, English, Polish, French, German, Serbian, and Croatian), we provide empirical evidence on the scale and structure of content reuse, reconceptualizing it as a core element of the global journalistic workflow.

\section{Our use case -- background on journalists' work}

As part of our research, we observed the work of journalists at the STA, is a national news organization reporting on domestic and international events. It produces news in two languages - Slovenian and English. In this research, we focus only on English news articles. 
We conducted interviews with journalists from STA and analyzed their editorial workflow. Focusing on the domestic perspective, we examined how STA journalists curate and adapt international news to align with the domestic media landscape. The dataset derived from this workflow is described in detail in Section \ref{sec:data}.
Slovenia has a comparatively limited representation of journalists abroad; therefore, its national news agency relies on content exchange agreements that provide access to international reports from regions without direct correspondents. Journalists covering foreign events aggregate and synthesize information from multiple agencies to produce domestically relevant news articles. Despite the large volume of available material, only a small subset is selected for publication based on perceived relevance to Slovenia and to the agency's subscribers. This selection process is guided by journalists’ implicit professional judgment rather than formalized or documented criteria, a judgment that derives from media industry experience, direct subscriber feedback, and indirect feedback based on analyses of article use and reuse in the Slovenian media environment.

\section{Literature Review}

\paragraph{Journalistic text reuse.}
Prior work in journalism has focused on building corpora and methods to study reuse and paraphrase in news. Notable resources include the METER (MEasuring TExt Reuse) corpus in English \citeplanguageresource{clough-etal-2002-building} and COUNTER for Urdu \citeplanguageresource{sharjeel_counter_2017}, which support evaluation of reuse in the press domain. Early methods relied on surface cues such as $n$-gram overlap, Greedy String Tiling, and sentence alignment \citeplanguageresource{clough_meter_2001}, while later work combined content, structure, and style features in classification models \citep{bar-etal-2012-text}. More recently, researchers have traced cross-language circulation of news (e.g., using BLAST to detect non-contiguous or editorially transformed reuse in Finnish historical journalism) \citep{Salmi02012021}. Taken together, these studies show that newsroom reuse is common and often involves edits and reorganization, but most work remains monolingual and does not model cross-language editorial pipelines or where reuse tends to appear within articles.

\paragraph{Intertextuality vs. editorial reuse.}
In literary studies, intertextuality refers to echoes and references within a single language, often involving deliberate signaling. \citet{burns-etal-2021-profiling} demonstrate that distributional models can surface non-lexical echoes (allusions) in a monolingual literary tradition of Latin epic poetry, providing new tools for literary criticism. Our setting, by contrast, concerns cross-document, cross-language editorial reuse driven by newsroom practice: editors select, paraphrase, summarize, and merge material from foreign sources, producing many-to-many sentence alignments under time pressure rather than deliberate literary signaling.

\paragraph{Translation-induced intertextuality.}
\citet{mcgovern-etal-2025-characterizing} examine how human and machine translation modulate intertextual links in Biblical texts, using multilingual embeddings to assess whether references are preserved or amplified. We differ in that we do not rely on a fixed catalog of known references. Instead, we automatically identify cross-language sentence correspondences as editors adapt foreign content, and we emphasize where such content appears in target articles over time. In short, we study how translation choices interact with newsroom workflows to shape cross-lingual reuse, rather than evaluating predefined links.

\paragraph{Rhetorical parallelism vs. cross-document alignment.}
\citet{bothwell-etal-2023-introducing} formalize rhetorical parallelism detection (RPD) as an intra-document sequence-labeling task that links stylistically parallel spans. While their focus is on within-text parallelism, the takeaway—that surface-level similarity can miss structural parallels—motivates our approach to cross-document, cross-lingual alignment, where paraphrase and restructuring are common and lexical overlap may be limited. Here, alignment refers to identifying sentences or passages with equivalent meaning across different news texts.

\paragraph{Cross-lingual alignment with multilingual embeddings.}
Multilingual sentence embeddings (e.g., multilingual SBERT) enable cross-language sentence retrieval without full translation, supporting scalable alignment across languages \citep{reimers-gurevych-2020-making}. We build on this capability but show that editorial reuse in journalism often involves many-to-many mappings and paraphrase or merging, which such embeddings alone struggle to capture. Our approach combines multilingual sentence representations with temporal filtering and discourse-aware analysis (e.g., where content sits within an article) to study not just whether reuse occurs, but how and where it is placed. Since the significance of news positioning has been established in various domains, 
including its effects on financial markets \cite{fedyk2018front}.

\section{Our Dataset and Approach}\label{sec:data}

\textbf{Dataset.} For this research, two datasets were created. STA provided us with the API, which allowed us to crawl news articles from the STA and foreign agencies. 
Thus, we collected a dataset of news articles produced by the STA for the period from October 7th, 2023, to November 2nd, 2023, and from February 1st, 2025, to February 28th, 2025. The first window covers the onset of the Israel-Hamas conflict. The second window lies outside the active conflict timeline, reflecting a comparatively calm period. We processed the STA dataset and excluded irrelevant news categories, such as weather reports and schedules of events, resulting in a final dataset of 1,037 news articles from the STA collection\footnote{Unfiltered STA dataset comprises 1,293 news articles.}. 
Furthermore, we collected a dataset of foreign articles for the same periods, provided to the STA by different foreign agencies (FA). The FA dataset comprises 237,551 news articles from 15 foreign agencies in 7 languages. The STA dataset contains 544 news articles from 2023 and 749 from 2025, while the FA dataset includes 120,668 articles from 2023 and 116,883 from 2025. The FA collection is multilingual, with articles written 84,119 in Italian, 49,545 in English, 26,364 in Polish, 25,482 in French, 24,353 in German, 18,378 in Serbian, and 9,310 in Croatian.  

Following, the datasets were cleaned from unimportant information, i.e., all HTML tags, emails, and telephone numbers were removed from news article texts in both datasets. 

\textbf{Weakly supervised method.} Based on our observations and conversations with journalists, we developed an approach designed to determine which foreign news articles were utilized to create a target STA news article. 
In the initial step, all the texts from the STA and the foreign corpus were segmented into sentences using the sentence-splitter library\footnote{\url{https://pypi.org/project/sentence-splitter/}}. 
Additionally, short sentences and sentences without verbs were excluded from the STA corpus\footnote{Tokenization and parsing parts of speech were conducted using the spaCy library.}. This decision was motivated by both linguistic theory and practical considerations related to the semantic representation of sentences.
Linguistic theories distinguish between content words (such as nouns, verbs, and adjectives) and function words (such as prepositions, conjunctions, or articles). The latter serve primarily grammatical purposes, and others contribute the most to the meaning of the sentence \citep{Katz1963-KATTSO-3, pustejovsky_lexical_1996}. Furthermore, in English, a meaningful sentence typically requires a verb. This contrasts with some other languages, e.g., those in the Slavic language family, where a meaningful sentence can be built without verbs. Additionally, we selected only sentences with more than 7 tokens to ensure minimal semantic complexity. Specifically, we retained only those sentences that contained adjectives, nouns, verbs, pronouns, proper nouns, and auxiliary verbs, which are crucial for capturing modality. 
This approach ensured that sentences such as those demonstrated in the following examples do not participate in the similarity mapping: 

\begin{itemize}
    \item \textit{Follow us also on:}
    \item \textit{Below is a schedule of events for Saturday, 1 February:}
    \item \textit{7:30am to 2pm: John Doe}
\end{itemize}

Subsequently, the sentences were transformed into vectors utilizing multilingual sentence transformers \cite{reimers-gurevych-2020-making}. 
Following, the comparison was restricted to news articles published on the same date. Using cosine similarity, each sentence in the STA news articles was compared with each sentence from the collection of FA news articles. Based on preliminary experiments conducted on the external dataset, we established a similarity score threshold of 60\%. 
The approach was evaluated on the Webis-Wikipedia-Text-Reuse-18 dataset \citeplanguageresource{alshomary_webis_2018}, which consists of documented text reuse cases from Wikipedia articles. From this corpus, we randomly sampled 1,095 text pairs. Given that our news dataset is multilingual, a portion of the selected pairs was translated into six target languages represented in our dataset using a multilingual translation model\footnote{\url{https://huggingface.co/facebook/nllb-200-distilled-600M}} \cite{team_no_2022}. The dataset statistics, test results, and detailed description of the approach are presented in Table~\ref{tab:webis_score} (Appendix~\ref{app:sim_test}).

The analysis showed that semantically similar sentences and full texts consistently yielded mean similarity scores above 0.6, whereas dissimilar sentence pairs did not exceed an average score of 0.4. 
Therefore, sentences exhibiting a cosine similarity greater than this threshold were marked as similar. The PyTorch implementation was used to calculate cosine similarity\footnote{\url{https://pytorch.org/docs/stable/generated/torch.nn.CosineSimilarity.html}}.

All articles in the STA dataset possess a timestamp, which indicates when an article was created. The timestamp includes year, month, day, hour, minute, second, and microsecond. All articles in the FA dataset also possess such a timestamp, indicating when the article was received by the STA. In the last preprocessing stage, the matched articles were flagged as false positives if the STA articles were created before the corresponding foreign articles were received\footnote{Detailed description of the approach and testing can be found at \url{https://github.com/kunturs/lrec2026-rewrite-news}.}.

\section{Results}\label{sec:results}

The similarity-based approach revealed 8,432 matched sentence pairs between STA (target) and FA (source). After excluding false positives, defined as matches found in STA articles published prior to the receipt of the corresponding FA articles, the number was reduced to 4,004. We further refined the dataset by retaining only the earliest matching FA article for each STA sentence, assuming it is the most likely origin of the content. This final filtering step delivered 1,087 matched sentence pairs\footnote{All the analysis was applied to the dataset containing only the earliest matches.}. Table~\ref{tab:sim_matches} demonstrates the number of matched sentences and news articles in each filtering step. 

\begin{table}[h]
\scriptsize
    \centering
    \begin{tabular}{p{.4\columnwidth}|p{.07\columnwidth}p{.09\columnwidth}|p{.09\columnwidth}p{.09\columnwidth}}
    \toprule
        & \multicolumn{2}{c|}{Articles} & \multicolumn{2}{c}{Sentences} \\
          & \rightline{STA} & \rightline{FA} & \rightline{STA} & \rightline{FA}\\
        \midrule
        True matches &  \rightline{387} & \rightline{2,070} & \rightline{989} & \rightline{2,731}\\
        The earliest matches & \centerline{{\,—}} & \rightline{698} & \centerline{{\,—}}  &  \rightline{875}\\
        False positives &  \rightline{419} & \rightline{2,376}  & \rightline{1,117}  & \rightline{3,288}\\
        \bottomrule
    \end{tabular}
    \caption{Number of sentences and articles matched using the similarity approach in the STA and FA datasets. An em dash (—) denotes “not applicable”. The earliest matches filtering only applies to the FA data.}
    \label{tab:sim_matches}
\end{table}

\begin{figure}[h]
  \centering

  \includegraphics[width=\linewidth]{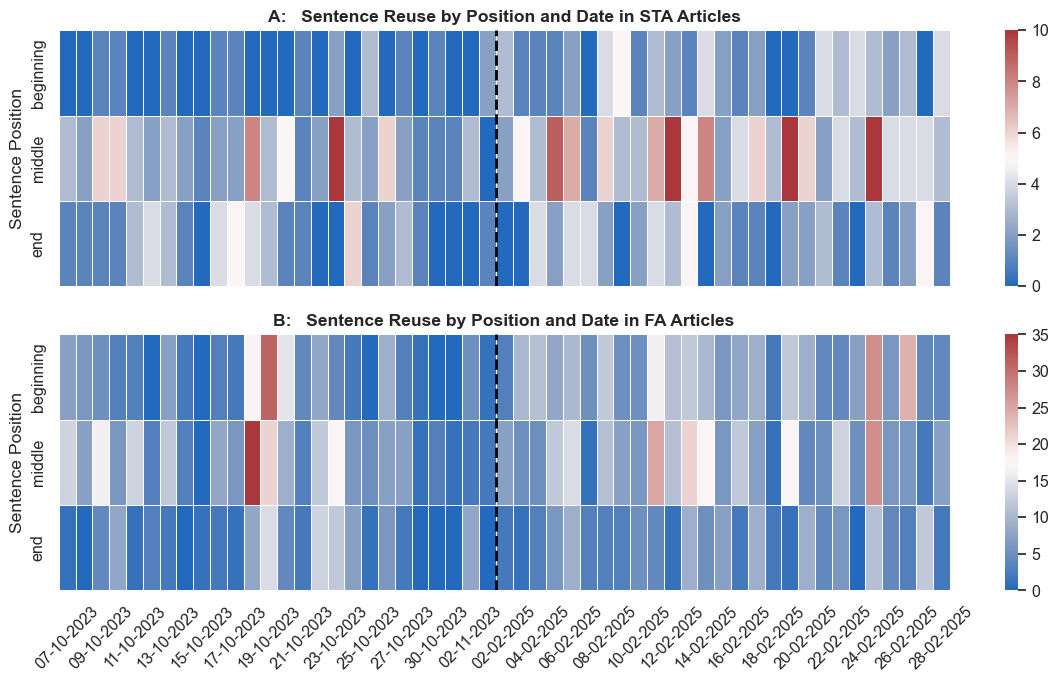}
    
  \caption{Distribution of reused sentences over time: (A) STA (sentences taken from FA) and (B) FA articles (sentences reused by STA). The plots show how reuse varies by sentence position (with a granularity of: beginning, middle, end of the article) within each article. The color scale on the right indicates the number of reused sentences: zero -- blue, through white up to red -- ten reused sentences. The dotted line marks the boundary between the years 2023 and 2025 in the data distribution.}
  \label{fig:reuse_heatmap}
\end{figure}

As shown in Figure~\ref{fig:reuse_heatmap}--A, sentence reuse is most prevalent in the middle sections of news articles, followed by the ending sections. In contrast, the beginning sections tend to contain the most original content. This pattern remains consistent over time. Similarly, the most reused content comes from the middle section of a news article, as Figure~\ref{fig:reuse_heatmap}-B demonstrates. However, more content are being reused from the beginning sections than from the ending sections. 
The analysis shows distinct positional reuse patterns between FA and STA. Most reused sentences occur from FA-middle to STA-middle (204 cases) and FA-beginning to STA-middle (217). Beginning-to-beginning reuse is infrequent (31). FA-middle sentences map to STA-beginning (130) and STA-end (144), while FA-beginning ones match with STA-end (135). FA-end sentences mainly abide near the end (112), with fewer mappings to the beginning (42) and middle (72). Overall, reuse concentrates in the middle sections of both article types.
The significance of the results was tested using the $\chi^2$ test of independence, resulting in a p-value less than the 0.05 threshold for all variables mentioned above. 

To investigate how textual reuse occurs across articles, sentence positional relationships (PRs) between STA and FA news sources were analyzed. A sentence positional relationship (PR) refers to how sentence(s) from an STA article align positionally with sentence(s) from an FA article that share semantically similar content. 
Each pair in the dataset matches one sentence from STA to one from FA. These PRs were categorized into four types based on sentence ID frequency: \textbf{1:1} (one-to-one match), \textbf{1:many} (one STA sentence aligned with multiple FA sentences), \textbf{many:1} (multiple STA sentences aligned with one FA sentence), and \textbf{many:many} (multiple PRs on both sides). The PR analysis was conducted in four steps. First, sentence pairs were filtered to ensure reuse was traced back to the earliest available foreign article. Then, sentence-level reuse frequency was counted, pairs were categorized into PR types, and the reuse direction was determined based on temporal order.
The majority of sentence reuse followed \textbf{many:many} (54.9\%) and \textbf{many:1} (40.5\%) structures, while direct one-to-one matches accounted for only 4.4\% of cases. These findings indicate that reuse is rarely literal or isolated; instead, it often involves paraphrasing, summarization, or segmentation across sentence boundaries.

\begin{figure}[h]
  \centering
  \includegraphics[width=\linewidth]{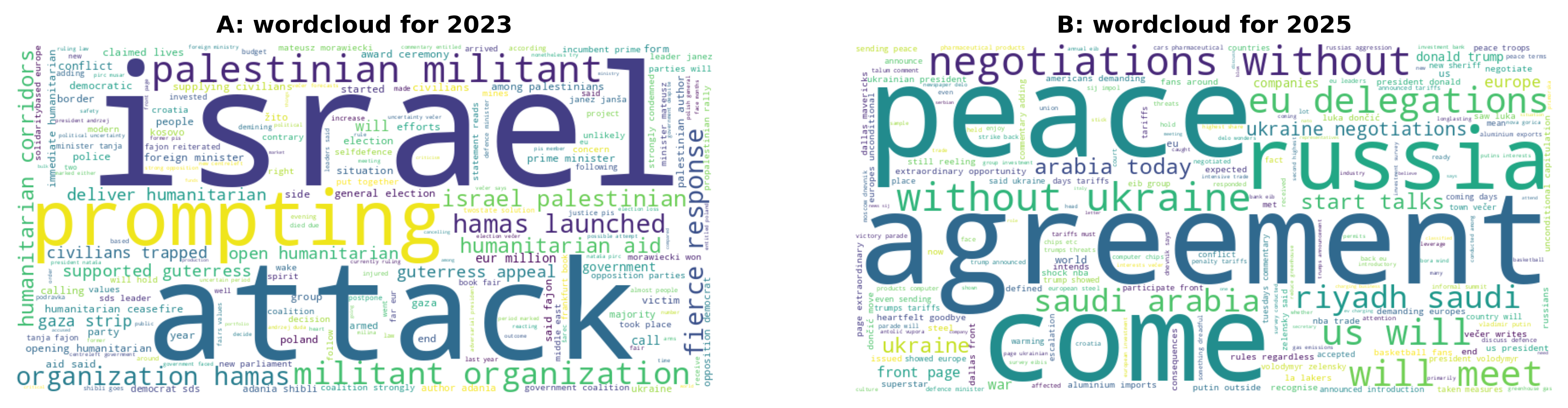}
  \caption{Word cloud illustrating recurrent lexical patterns in reused sentences, highlighting the shared editorial templates underlying coverage during major news cycles: corresponding to the most frequently reused sentences in (A)  2023 and in (B) 2025.}
  \label{fig:wordcloud}
\end{figure}

Qualitative analysis of the most frequently reused sentences (corresponding to the red peaks in Figure~\ref{fig:reuse_heatmap}-A) showed that peaks in sentence reuse in the years 2023 (Figure~\ref{fig:wordcloud}-A) and 2025 (Figure~\ref{fig:wordcloud}-B) correspond to intense news cycles dominated, respectively, by a national crisis and by international political reporting. In both cases, repetition resulted not from copying but from the systematic use of shared editorial templates to cover closely related events.

\section{Conclusion}
In this research, we investigate sentence-level text reuse in multilingual journalism. We propose a weakly supervised framework to detect reuse between English-language articles from the STA and 15 international news agencies publishing in seven languages. Our analysis indicates that 52\% of STA news and approximately 1.6\% of news articles from international agencies share similar content. Our findings suggest that passages originating from the beginning and middle sections of news articles are reused more frequently than those from the concluding sections. Conversely, the reused content tends to be positioned primarily in the middle and at the end of news articles, suggesting that the beginning sections are more likely to contain original material. 

The distribution of PR types suggests that textual reuse between STA and FA news sources follows editorial practices that involve restructuring rather than direct copying. The low proportion of one-to-one PRs indicates that literal reuse is not the dominant strategy. Instead, the prevalence of many-to-many and many-to-one cases points to more flexible reuse patterns, likely shaped by summarization, paraphrasing, and contextual adaptation. 
These findings reveal the limitations of surface-level similarity approaches and emphasize the need for discourse-aware, multilingual NLP tools. By situating NLP methods within journalistic workflows, the study offers new directions for modeling cross-lingual information flow and encourages interdisciplinary collaboration across language technology, media studies, and computational social science.

\section{Limitations}\label{sec:limitations}

The primary challenge in developing the methodology for detecting text reuse originated from working with a dataset containing multilingual data in seven languages. Given the computational costs and time constraints, we opted against translating all articles into English. Instead, we utilized a multilingual language representation model \cite{reimers-gurevych-2020-making} and libraries capable of handling data in multiple languages.

In addition to the Webis-Wikipedia-Text-Reuse-18 dataset \citeplanguageresource{alshomary_webis_2018}, we evaluated our method on the Webis-CPC-11 dataset \citeplanguageresource{burrows_webis_2013}. These corpora were chosen due to their open availability. Webis-CPC-11  contains annotated paraphrased and non-paraphrased sentences. 
Due to the computational and processing time, for our tests, we used only part of both datasets. 
Specifically, from the Webis-CPC-11 corpus, a random sample of 1,000 text pairs was drawn, and our similarity approach was applied to this selection. The testing datasets, along with a detailed description of the approach and testing, can be found in the GitHub repository\footnote{ \url{https://github.com/kunturs/lrec2026-rewrite-news}.}.
Additionally, the dataset statistics, test results, and description of the approach are presented in Table~\ref{tab:webis_score} (Appendix~\ref{app:sim_test}). 
The mean similarity score for both paraphrased and non-paraphrased samples exceeds 0.7, implying that the current approach is ineffective for distinguishing paraphrased from non-paraphrased text. Therefore, this approach cannot be applied to differentiate the stance through which the information was conveyed.  This high similarity score may be attributed to the nature of the corpus, in which non-paraphrased sentence pairs, despite differing in final meaning, often remain semantically similar, as illustrated by the example below: 
\begin{itemize}
    \item \textbf{Original text:} \textit{"I have heard many accounts of him," said Emily, "all differing from each other: I think, however, that the generality of people rather incline to Mrs. Dalton\'s opinion than to yours, Lady Margaret."  "I can easily believe it.'}
    \item \textbf{Non-paraphrased text:} \textit{"I have heard many accounts of him," said Emily, "all different from each other: I think, however, that the generality of the people rather inclined to the view of Ms Dalton to yours, Lady Margaret." That I can not believe.}
\end{itemize}

We opted for using cosine similarity against other similarity measures, as this approach is commonly used in many NLP tasks, is computationally efficient, and easily interpretable. 

Some frequent words were deliberately deleted from Figure~\ref{fig:wordcloud} due to anonymity issues. The deleted words will be displayed in the accepted version of the paper.

\section{Acknowledgements}

All authors were funded by the European Union under the Horizon Europe grant OMINO (grant number 101086321, \cite{holyst2024}). Views and opinions expressed are, however, those of the authors
only and do not necessarily reflect those of the European Union or the European Research Executive Agency. Neither the European Union nor the European Research Executive Agency can be held responsible for them. A.W. and S.K. were also co-financed with funds from the Polish Ministry of Education and Science under the program entitled International Co-Financed Projects. Nina Smirnova additionally received funding from the Deutsche Forschungsgemeinschaft (DFG) under grant number: MA 3964/7-3 (POLLUX Project).

\section{Bibliographical References}\label{sec:reference}

\bibliographystyle{lrec2026-natbib}
\bibliography{lrec2026-example}

\begin{thebibliography}{4}
\expandafter\ifx\csname natexlab\endcsname\relax\def\natexlab#1{#1}\fi

\bibitem[{Alshomary et~al.(2018)Alshomary, Völske, Wachsmuth, Stein, Hagen, and Potthast}]{alshomary_webis_2018}
Alshomary, Milad and Völske, Michael and Wachsmuth, Henning and Stein, Benno and Hagen, Matthias and Potthast, Martin. 2018.
\newblock \href {https://doi.org/10.5281/ZENODO.3546193} {\emph{Webis Wikipedia Text Reuse Corpus 2018 (Webis-Wikipedia-Text-Reuse-18)}}.
\newblock Zenodo.

\bibitem[{Burrows et~al.(2013)Burrows, Potthast, Stein, and Eiselt}]{burrows_webis_2013}
Burrows, Steven and Potthast, Martin and Stein, Benno and Eiselt, Andreas. 2013.
\newblock \href {https://doi.org/10.5281/ZENODO.3251771} {\emph{Webis Crowd Paraphrase Corpus 2011 (Webis-{CPC}-11)}}.
\newblock Zenodo.

\bibitem[{Clough et~al.(2001)Clough, Gaizauskas, Piao, and Wilks}]{clough_meter_2001}
Clough, Paul and Gaizauskas, Robert and Piao, Scott S. L. and Wilks, Yorick. 2001.
\newblock \href {https://doi.org/10.3115/1073083.1073110} {\emph{{METER}: {MEasuring} {TExt} Reuse}}.
\newblock Association for Computational Linguistics.

\bibitem[{Sharjeel et~al.(2017)Sharjeel, Nawab, and Rayson}]{sharjeel_counter_2017}
Sharjeel, Muhammad and Nawab, Rao Muhammad Adeel and Rayson, Paul. 2017.
\newblock \href {https://doi.org/10.1007/s10579-016-9367-2} {\emph{{COUNTER}: corpus of Urdu news text reuse}}.

\end{thebibliography}


\begin{thebibliography}{14}
\expandafter\ifx\csname natexlab\endcsname\relax\def\natexlab#1{#1}\fi

\bibitem[{B{\"a}r et~al.(2012)B{\"a}r, Zesch, and Gurevych}]{bar-etal-2012-text}
Daniel B{\"a}r, Torsten Zesch, and Iryna Gurevych. 2012.
\newblock \href {https://aclanthology.org/C12-1011/} {Text reuse detection using a composition of text similarity measures}.
\newblock In \emph{Proceedings of {COLING} 2012}, pages 167--184, Mumbai, India. The COLING 2012 Organizing Committee.

\bibitem[{Bothwell et~al.(2023)Bothwell, DeBenedetto, Crnkovich, M{\"u}ller, and Chiang}]{bothwell-etal-2023-introducing}
Stephen Bothwell, Justin DeBenedetto, Theresa Crnkovich, Hildegund M{\"u}ller, and David Chiang. 2023.
\newblock \href {https://doi.org/10.18653/v1/2023.emnlp-main.305} {Introducing rhetorical parallelism detection: A new task with datasets, metrics, and baselines}.
\newblock In \emph{Proceedings of the 2023 Conference on Empirical Methods in Natural Language Processing}, pages 5007--5039, Singapore. Association for Computational Linguistics.

\bibitem[{Burns et~al.(2021)Burns, Brofos, Li, Chaudhuri, and Dexter}]{burns-etal-2021-profiling}
Patrick~J. Burns, James~A. Brofos, Kyle Li, Pramit Chaudhuri, and Joseph~P. Dexter. 2021.
\newblock \href {https://doi.org/10.18653/v1/2021.naacl-main.389} {Profiling of intertextuality in {L}atin literature using word embeddings}.
\newblock In \emph{Proceedings of the 2021 Conference of the North American Chapter of the Association for Computational Linguistics: Human Language Technologies}, pages 4900--4907, Online. Association for Computational Linguistics.

\bibitem[{Clough et~al.(2002)Clough, Gaizauskas, and Piao}]{clough-etal-2002-building}
Paul Clough, Robert Gaizauskas, and S.~L. Piao. 2002.
\newblock \href {https://aclanthology.org/L02-1218/} {Building and annotating a corpus for the study of journalistic text reuse}.
\newblock In \emph{Proceedings of the Third International Conference on Language Resources and Evaluation ({LREC}`02)}, Las Palmas, Canary Islands - Spain. European Language Resources Association (ELRA).

\bibitem[{Costa-jussà et~al.(2022)Costa-jussà, Cross, Çelebi, Elbayad, Heafield, Heffernan, Kalbassi, Lam, Licht, Maillard, Sun, Wang, Wenzek, Youngblood, Akula, Barrault, Gonzalez, Hansanti, Hoffman, Jarrett, Sadagopan, Rowe, Spruit, Tran, Andrews, Ayan, Bhosale, Edunov, Fan, Gao, Goswami, Guzmán, Koehn, Mourachko, Ropers, Saleem, Schwenk, and Wang}]{team_no_2022}
Marta~R. Costa-jussà, James Cross, Onur Çelebi, Maha Elbayad, Kenneth Heafield, Kevin Heffernan, Elahe Kalbassi, Janice Lam, Daniel Licht, Jean Maillard, Anna Sun, Skyler Wang, Guillaume Wenzek, Al~Youngblood, Bapi Akula, Loic Barrault, Gabriel~Mejia Gonzalez, Prangthip Hansanti, John Hoffman, Semarley Jarrett, Kaushik~Ram Sadagopan, Dirk Rowe, Shannon Spruit, Chau Tran, Pierre Andrews, Necip~Fazil Ayan, Shruti Bhosale, Sergey Edunov, Angela Fan, Cynthia Gao, Vedanuj Goswami, Francisco Guzmán, Philipp Koehn, Alexandre Mourachko, Christophe Ropers, Safiyyah Saleem, Holger Schwenk, and Jeff Wang. 2022.
\newblock \href {https://doi.org/10.48550/arXiv.2207.04672} {No language left behind: Scaling human-centered machine translation}.

\bibitem[{D{'}hondt et~al.(2016)D{'}hondt, Grouin, and Grau}]{dhondt-etal-2016-low}
Eva D{'}hondt, Cyril Grouin, and Brigitte Grau. 2016.
\newblock \href {https://doi.org/10.18653/v1/W16-6108} {Low-resource {OCR} error detection and correction in {F}rench clinical texts}.
\newblock In \emph{Proceedings of the Seventh International Workshop on Health Text Mining and Information Analysis}, pages 61--68, Auxtin, TX. Association for Computational Linguistics.

\bibitem[{Fedyk(2024)}]{fedyk2018front}
Anastassia Fedyk. 2024.
\newblock \href {https://doi.org/10.1111/jofi.13287} {Front‐page news: The effect of news positioning on financial markets}.
\newblock \emph{The Journal of Finance}, 79(1):5--33.

\bibitem[{Hołyst et~al.(2024)Hołyst, Mayr, Thelwall, Frommholz, Havlin, Sela, Kenett, Helic, Rehar, Maček, Kazienko, Kajdanowicz, Biecek, Szymanski, and Sienkiewicz}]{holyst2024}
Janusz~A. Hołyst, Philipp Mayr, Michael Thelwall, Ingo Frommholz, Shlomo Havlin, Alon Sela, Yoed~N. Kenett, Denis Helic, Aljoša Rehar, Sebastijan~R. Maček, Przemysław Kazienko, Tomasz Kajdanowicz, Przemysław Biecek, Boleslaw~K. Szymanski, and Julian Sienkiewicz. 2024.
\newblock \href {https://doi.org/10.1038/s41562-024-01833-8} {Protect our environment from information overload}.
\newblock \emph{Nature Human Behaviour}, 8:402--403.

\bibitem[{Katz and Fodor(1963)}]{Katz1963-KATTSO-3}
Jerrold Katz and Jerry Fodor. 1963.
\newblock The structure of a semantic theory.
\newblock \emph{Language}, 39:170--210.

\bibitem[{Lee(2007)}]{lee-2007-computational}
John Lee. 2007.
\newblock \href {https://aclanthology.org/P07-1060/} {A computational model of text reuse in ancient literary texts}.
\newblock In \emph{Proceedings of the 45th Annual Meeting of the Association of Computational Linguistics}, pages 472--479, Prague, Czech Republic. Association for Computational Linguistics.

\bibitem[{McGovern et~al.(2025)McGovern, Sirin, and Lippincott}]{mcgovern-etal-2025-characterizing}
Hope McGovern, Hale Sirin, and Tom Lippincott. 2025.
\newblock \href {https://doi.org/10.18653/v1/2025.naacl-short.14} {Characterizing the effects of translation on intertextuality using multilingual embedding spaces}.
\newblock In \emph{Proceedings of the 2025 Conference of the Nations of the Americas Chapter of the Association for Computational Linguistics: Human Language Technologies (Volume 2: Short Papers)}, pages 161--167, Albuquerque, New Mexico. Association for Computational Linguistics.

\bibitem[{Pustejovsky(1996)}]{pustejovsky_lexical_1996}
James Pustejovsky. 1996.
\newblock Lexical semantics: the problem of polysemy.

\bibitem[{Reimers and Gurevych(2020)}]{reimers-gurevych-2020-making}
Nils Reimers and Iryna Gurevych. 2020.
\newblock \href {https://doi.org/10.18653/v1/2020.emnlp-main.365} {Making monolingual sentence embeddings multilingual using knowledge distillation}.
\newblock In \emph{Proceedings of the 2020 Conference on Empirical Methods in Natural Language Processing (EMNLP)}, pages 4512--4525, Online. Association for Computational Linguistics.

\bibitem[{Salmi et~al.(2021)Salmi, Paju, Rantala, Nivala, Vesanto, and Ginter}]{Salmi02012021}
Hannu Salmi, Petri Paju, Heli Rantala, Asko Nivala, Aleksi Vesanto, and Filip Ginter. 2021.
\newblock \href {https://doi.org/10.1080/01615440.2020.1803166} {The reuse of texts in finnish newspapers and journals, 1771–1920: A digital humanities perspective}.
\newblock \emph{Historical Methods: A Journal of Quantitative and Interdisciplinary History}, 54(1):14--28.

\end{thebibliography}

\section{Language Resource References}
\label{lr:ref}
\bibliographystylelanguageresource{lrec2026-natbib}
\bibliographylanguageresource{languageresource}

\clearpage
\appendix
\onecolumn

\section{Testing of the similarity approach}\label{app:sim_test}

To establish similarity thresholds for both sentence-level and document-level approaches, we utilized the Webis-Wikipedia-Text-Reuse-18 corpus, which contains pairs of short texts and their reused counterparts. Each text pair is first segmented into sentences, yielding aligned source (\texttt{source text}) and target (\texttt{target text}) sentence sequences, as Table~\ref{tab:wiki_example} demonstrates. We assumed a monotonic alignment, i.e., that sentences correspond sequentially between source and target.

\begin{table}[H]
\small
      \centering
\begin{tabular}{p{.45\textwidth}p{.45\textwidth}}
\toprule
source text & target text \\
\midrule
According to the United States Census Bureau, the CDP has a total area of , of which, of it is land and of it (0.05 ) is water. & According to the United States Census Bureau, the village has a total area of 2.8 square miles (7.1 km ), of which, 2.4 square miles (6.2 km ) of it is land and 0.3 square miles (0.9 km ) of it (12.32 ) is water. \\
\midrule
Wiseman has a continental subarctic climate (Köppen Dfc). & Babylon Village is bordered to the west by West Babylon, to the north by North Babylon, to the east by West Islip, and to the south by the Great South Bay. \\
\midrule
As of the census of 2000, there were 21 people, 7 households, and 3 families residing in the CDP. & As of the census of 2010, there were 12,166 people and 4,585 households in the village, with 2.65 persons per household. \\
\midrule
The population density was 0.3 people per square mile (0.1 km ). & The population density was 4,975.9 people per square mile (2,864.3 km ). \\
\midrule
There were 30 housing units at an average density of 0.4 sq mi (0.1 km ). & There were 4,768 housing units. \\
\bottomrule
\end{tabular}
 \caption{Example of the sentence alignment in the Webis-Wikipedia-Text-Reuse-18 corpus.}
   \label{tab:wiki_example}
\end{table}

For each text pair, we compute similarity at two levels. First, we obtain a document-level similarity score by encoding the full source and target texts (concatenated sentences) using a sentence embedding model and computing their cosine similarity.

Second, we compute sentence-level similarities by encoding each sentence individually. We then perform pairwise comparisons between all source and target sentence embeddings. Based on the sequential alignment assumption, similarities are divided into two categories: (i) aligned sentence pairs, where source and target sentences share the same index, and (ii) non-aligned sentence pairs, where indices differ. For each category, we calculate the mean cosine similarity across all comparisons.

This procedure yields three scores per text pair: (1) average full-text similarity, (2) average similarity of non-aligned sentences (representing unrelated content), and (3) average similarity of aligned sentences (representing reused content), as Table~\ref{tab:wiki_scores} shows. These aggregated scores are then used to analyze the separation between reused and non-reused text segments and to derive empirical similarity thresholds.

\begin{table}[H]
      \centering
\begin{tabular} 
{p{.1\textwidth}p{.15\textwidth}p{.15\textwidth}p{.15\textwidth}p{.15\textwidth}}
\toprule
 language & full text & different \newline sentences &  similar \newline sentences & support\\
\midrule
English & 0.72 & 0.37 & 0.63 & 500 \\
German & 0.73 & 0.35 & 0.64 & 99 \\
French & 0.69 & 0.35 & 0.59 & 99 \\
Croatian & 0.71 & 0.37 & 0.63 & 99 \\
Italian & 0.74 & 0.37 & 0.69 & 100 \\
Polish & 0.69 & 0.35 & 0.63 & 99 \\
Serbian & 0.72 & 0.37 & 0.69 & 99 \\
\bottomrule
\end{tabular}
 \caption{Similarity scores for the Webis-Wikipedia-Text-Reuse-18 corpus.}
   \label{tab:wiki_scores}
\end{table}

The same procedure was applied to the Webis-CPC-11 corpus, which contains both paraphrased and non-paraphrased versions of original texts. Table~\ref{tab:webis_score} reports the average full-text similarity scores. We restrict our analysis to document-level similarity only in English, as the results indicate that the proposed method is not sufficiently effective for paraphrase detection at this level. Consequently, we do not extend this approach to sentence-level analysis.

\begin{table}[H]
    \centering
    \begin{tabular}{p{.3\textwidth}p{.2\textwidth}p{.15\textwidth}}
    \toprule
        paraphrased & full text & support\\
    \midrule
         no & 0.78 & 500\\
         yes & 0.81 & 500 \\
\bottomrule
    \end{tabular}
    \caption{Similarity scores for the Webis-CPC-11}
 \label{tab:webis_score}
\end{table}

\end{document}